\documentclass[letterpaper]{article} 
\usepackage{aaai23}  
\usepackage{times}  
\usepackage{helvet}  
\usepackage{courier}  
\usepackage[hyphens]{url}  
\usepackage{graphicx} 
\usepackage{natbib}  
\usepackage{caption} 
\usepackage{algorithm}
\usepackage{algorithmic}

\usepackage[utf8]{inputenc} 
\usepackage[T1]{fontenc}    
\usepackage{hyperref}       
\usepackage{url}            
\usepackage{booktabs}       
\usepackage{amsfonts}       
\usepackage{nicefrac}       
\usepackage{microtype}      
\usepackage{xcolor}         

\hypersetup{
    colorlinks,
    linkcolor={red!50!black},
    citecolor={blue!50!black},
    urlcolor={blue!80!black}
}

\usepackage{graphicx}
\usepackage{subfigure}
\usepackage{todonotes}

\newcommand{\hlc}[2]{{\color{#1}#2}}

\newcommand{\del}[1]{}
\newcommand{\ins}[1]{\hlc{black}{#1}} 

\usepackage{amssymb} 

\usepackage{amsmath}
\usepackage{siunitx}
\usepackage{listings}

\newcommand{\secInLine}[1]{\textbf{#1}.}

\usepackage{acro} 
\DeclareAcronym{milp}{
  short=MILP,
  long=mixed integer linear programme,
}

\DeclareAcronym{bnb}{
  short=B\&B,
  long=branch-and-bound,
}

\DeclareAcronym{sota}{
  short=SOTA,
  long=state-of-the-art,
}

\DeclareAcronym{rl}{
  short=RL,
  long=reinforcement learning,
}

\DeclareAcronym{dqb}{
  short=DQB,
  long=deep Q-branching,
}

\DeclareAcronym{co}{
  short=CO,
  long=combinatorial optimisation,
}

\DeclareAcronym{ml}{
  short=ML,
  long=machine learning,
}

\DeclareAcronym{dnn}{
  short=DNN,
  long=deep neural network,
}

\DeclareAcronym{gnn}{
  short=GNN,
  long=graph neural network,
}

\DeclareAcronym{sb}{
  short=SB,
  long=strong branching,
}

\DeclareAcronym{mdp}{
  short=MDP,
  long=Markov decision process,
}

\DeclareAcronym{dqfd}{
  short=DQfD,
  long=deep Q-learning from demonstrations,
}

\DeclareAcronym{mib}{
  short=MIB,
  long=most infeasible branching,
}

\DeclareAcronym{pb}{
  short=PB,
  long=pseudocost branching,
}

\DeclareAcronym{svm}{
  short=SVM,
  long=support vector machine,
}

\DeclareAcronym{gcn}{
  short=GCN,
  long=graph convolutional network,
}

\DeclareAcronym{fmsts}{
  short=FMSTS,
  long=fitting for minimising the sub-tree size,
}

\DeclareAcronym{dqn}{
  short=DQN,
  long=deep Q-network,
}

\DeclareAcronym{lp}{
  short=LP,
  long=linear programme,
}

\DeclareAcronym{il}{
  short=IL,
  long=imitation learning,
}

\DeclareAcronym{rpb}{
  short=RPB,
  long=reliability pseudocost branching,
}

\DeclareAcronym{bfs}{
  short=BFS,
  long=breadth-first search,
}

\DeclareAcronym{dfs}{
  short=DFS,
  long=depth-first search,
}

\usepackage{newfloat}
\usepackage{listings}
\DeclareCaptionStyle{ruled}{labelfont=normalfont,labelsep=colon,strut=off} 
\floatstyle{ruled}
\newfloat{listing}{tb}{lst}{}
\floatname{listing}{Listing}
\title{Reinforcement Learning for Branch-and-Bound Optimisation\\using Retrospective Trajectories}
\author{
    Christopher W. F. Parsonson,\textsuperscript{\rm 1}\thanks{Work undertaken during internship at InstaDeep.\\Corresponding email: cwfparsonson@gmail.com}
    Alexandre Laterre,\textsuperscript{\rm 2}
    Thomas D. Barrett\textsuperscript{\rm 2}
}
\affiliations{
    \textsuperscript{\rm 1}UCL, 
    \textsuperscript{\rm 2}InstaDeep\\

}

\usepackage{bibentry}

\begin{document}

\maketitle




\begin{abstract}

Combinatorial optimisation problems framed as mixed integer linear programmes (MILPs) are ubiquitous across a range of real-world applications. The canonical branch-and-bound algorithm seeks to exactly solve MILPs by constructing a search tree of increasingly constrained sub-problems. In practice, its solving time performance is dependent on heuristics, such as the choice of the next variable to constrain (`branching'). Recently, machine learning (ML) has emerged as a promising paradigm for branching. However, prior works have struggled to apply reinforcement learning (RL), citing sparse rewards, difficult exploration, and partial observability as significant challenges. Instead, leading ML methodologies resort to approximating high quality handcrafted heuristics with imitation learning (IL), which precludes the discovery of novel policies and requires expensive data labelling. In this work, we propose \textit{retro branching}; a simple yet effective approach to RL for branching. By retrospectively deconstructing the search tree into multiple paths each contained within a sub-tree, we enable the agent to learn from shorter trajectories with more predictable next states. In experiments on four combinatorial tasks, our approach enables learning-to-branch without any expert guidance or pre-training. We outperform the current state-of-the-art RL branching algorithm by $3$-$5\times$ and come within $20\%$ of the best IL method's performance on MILPs with \num{500} constraints and \num{1000} variables, with ablations verifying that our retrospectively constructed trajectories are essential to achieving these results.

\end{abstract}

\section{Introduction}
\label{sec:introduction}


A plethora of real-world problems fall under the broad category of \ac{co} (vehicle routing and scheduling \citep{korte12}; protein folding \citep{perdomo12}; fundamental science \citep{barahona82}). Many \ac{co} problems can be formulated as \acp{milp} whose task is to assign discrete values to a set of decision variables, subject to a mix of linear and integrality constraints, such that some objective function is maximised or minimised. The most popular method for finding exact solutions to \acp{milp} is \ac{bnb} \citep{land1960}; a collection of heuristics which increasingly tighten the bounds in which an optimal solution can reside (see Section \ref{sec:background}). Among the most important of these heuristics is \textit{variable selection} or \textit{branching} (which variable to use to partition the chosen node's search space), which is key to determining \ac{bnb} solve efficiency \citep{Achterberg2013}.

\begin{figure*}[!tp]
    \centering
    \includegraphics[width=\textwidth]{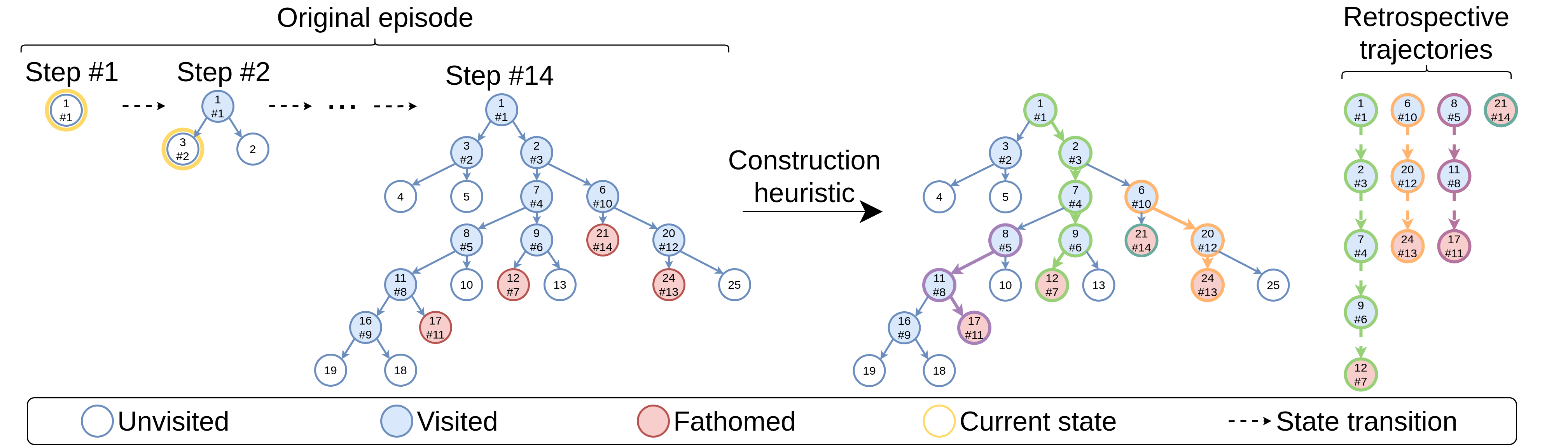}
    \caption{The proposed retro branching approach used during training. Each node is labelled with: Top: The unique ID assigned when it was added to the tree, and (where applicable); bottom: The step number (preceded by a `\#') at which it was visited by the brancher in the original \ac{mdp}. The \acs{milp} is first solved with the brancher and the \ac{bnb} tree stored as usual (forming the `original episode'). Then, ignoring any nodes never visited by the agent, the nodes are added to trajectories using some `construction heuristic' (see Sections \ref{sec:methodology} and \ref{sec:results_and_discussion}) until each eligible node has been added to one, and only one, trajectory. Crucially, the order of the sequential states within a given trajectory may differ from the state visitation order of the original episode, but all states within the trajectory will be within the same sub-tree. These trajectories are then used for training.}
    \label{fig:subtree_episodes_construction}
\end{figure*}

\Ac{sota} learning-to-branch approaches typically use the \ac{il} paradigm to predict the action of a high quality but computationally expensive human-designed branching expert \citep{gasse2019}. Since branching can be formulated as a \ac{mdp} \citep{He2014}, \ac{rl} seems a natural approach. The long-term motivations of \Ac{rl} include the promise of learning novel policies from scratch without the need for expensive expert data, the potential to exceed expert performance without human design, and the capability to maximise the performance of a policy parameterised by an expressivity-constrained \ac{dnn}.

However, branching has thus far proved largely intractable for \ac{rl} for reasons we summarise into three key challenges. (1) \textit{Long episodes:} Whilst even random branching policies are theoretically guaranteed to eventually find the optimal solution, poor decisions can result in episodes of tens of thousands of steps for the \num{500} constraint \num{1000} variable \acp{milp} considered by \citealt{gasse2019}. This raises the familiar \ac{rl} challenges of reward sparsity \citep{Trott2019KeepingYD}, credit assignment \citep{Harutyunyan2019}, and high variance returns \citep{Mao2019}. (2) \textit{Large state-action spaces}: Each branching step might have hundreds or thousands of potential branching candidates with a huge number of unique possible sub-\ac{milp} states. Efficient exploration to discover improved trajectories in such large state-action spaces is a well-known difficulty for \ac{rl} \citep{Agostinelli2019, ecoffet2021goexplore}. (3) \textit{Partial observability}: When a branching decision is made, the next state given to the brancher is determined by the next sub-\ac{milp} visited by the node selection policy. Jumping around the \ac{bnb} tree without the brancher's control whilst having only partial observability of the full tree makes the future states seen by the agent difficult to predict. \citealt{etheve2020} therefore postulated the benefit of keeping the \ac{mdp} within a sub-tree to improve observability and introduced the \ac{sota} \ac{fmsts} \ac{rl} branching algorithm. However, in order to achieve this, \ac{fmsts} had to use a \ac{dfs} node selection policy which, as we demonstrate in Section \ref{sec:results_and_discussion}, is highly sub-optimal and limits scalability.

In this work, we present \textit{retro branching}; a simple yet effective method to overcome the above challenges and learn to branch via reinforcement. We follow the intuition of \citet{etheve2020} that constraining each sequential \ac{mdp} state to be within the same sub-tree will lead to improved observability. However, we posit that a branching policy taking the `best' actions with respect to only the sub-tree in focus can still provide strong overall performance \textit{regardless of the node selection policy used}. This is aligned with the observation that leading heuristics such as SB and PB also do not explicitly account for the node selection policy or predict how the global bound may change as a result of activity in other sub-trees. Assuming the validity of this hypothesis, we can discard the \ac{dfs} node selection requirement of \ac{fmsts} whilst retaining the condition that sequential states seen during training must be within the same sub-tree.

Concretely, our retro branching approach (shown in Figure \ref{fig:subtree_episodes_construction} and elaborated on in Section \ref{sec:methodology}) is to, during training, take the search tree after the \ac{bnb} instance has been solved and \textit{retrospectively} select each subsequent state (node) to construct multiple trajectories. Each trajectory consists of sequential nodes within a single sub-tree, allowing the brancher to learn from shorter trajectories with lower return variance and more predictable future states. This approach directly addresses challenges (1) and (3) and, whilst the state-action space is still large, the shorter trajectories implicitly define more immediate auxiliary objectives relative to the tree. This reduces the difficulty of exploration since shorter trajectory returns will have a higher probability of being improved upon via stochastic action sampling than when a single long \ac{mdp} is considered, thereby addressing (2). Furthermore, retro branching relieves the \ac{fmsts} requirement that the agent must be trained in a \ac{dfs} node selection setting, enabling more sophisticated strategies to be used which are better suited for solving larger, more complex \acp{milp}.

We evaluate our approach on \acp{milp} with up to \num{500} constraints and \num{1000} variables, achieving a $3$-$5\times$ improvement over \ac{fmsts} and coming within $\approx20\%$ of the performance of the \ac{sota} \ac{il} agent of \citet{gasse2019}. Furthermore, we demonstrate that, for small instances, retro branching can uncover policies superior to \ac{il}; a key motivation of using \ac{rl}. Our results open the door to the discovery of new branching policies which can scale without the need for labelled data and which could, in principle, exceed the performance of \ac{sota} handcrafted branching heuristics.

\section{Related Work}
\label{sec:related_work}

Since the invention of \ac{bnb} for exact \ac{co} by \citet{land1960}, researchers have sought to design and improve the node selection (tree search), variable selection (branching), primal assignment, and pruning heuristics used by \ac{bnb}, with comprehensive reviews provided by \citealt{achterberg2007} and \citealt{Tomazella2020}. We focus on branching.

\secInLine{Classical branching heuristics} 
\Ac{pb} \citep{Benichou1971} and \ac{sb} \citep{Applegate1995, Applegate2007} are two canonical branching algorithms. \Ac{pb} selects variables based on their historic branching success according to metrics such as bound improvement. Although the per-step decisions of \ac{pb} are computationally fast, it must initialise the variable pseudocosts in some way which, if done poorly, can be particularly damaging to overall performance since early \ac{bnb} decisions tend to be the most influential. \Ac{sb}, on the other hand, conducts a one-step lookahead for all branching candidates by computing their potential local dual bound gains before selecting the most favourable variable, and thus is able to make high quality decisions during the critical early stages of the search tree's evolution. Despite its simplicity, \ac{sb} is still today the best known policy for minimising the overall number of \ac{bnb} nodes needed to solve the problem instance (a popular \ac{bnb} quality indicator). However, its computational cost renders \ac{sb} infeasible in practice.

\begin{figure*}[!tp]
    \centering
    \includegraphics[width=\textwidth, clip]{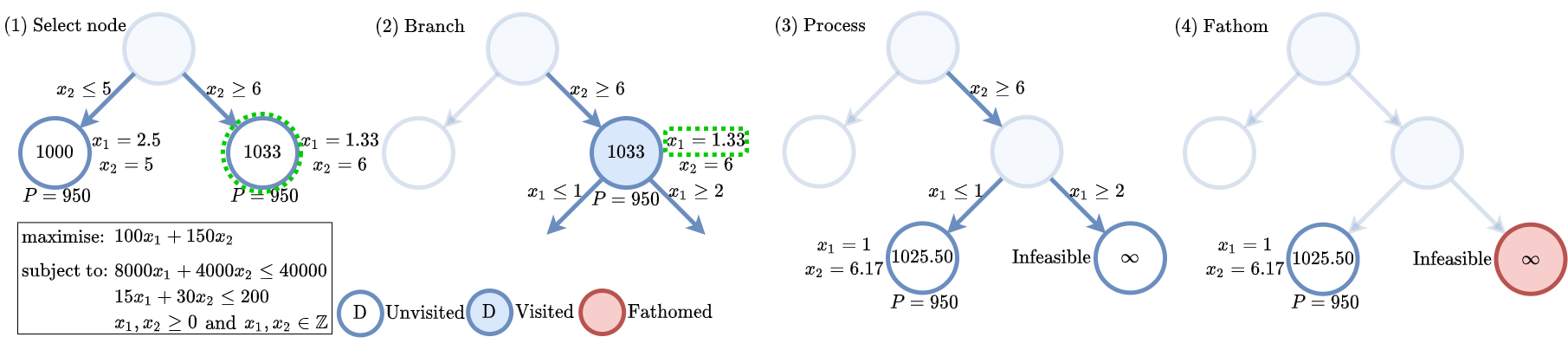}
    \caption{Typical $4$-stage procedure iteratively repeated by \ac{bnb} to solve an \ac{milp}. Each node represents an \ac{milp} derived from the original \ac{milp} being solved, and each edge represents the constraint added to derive a new child node (sub-\ac{milp}) from a given parent. Each node is labelled with the decision variable values of the solved LP relaxation on the right hand side, the corresponding dual bound in the centre, and the established primal bound beneath. Each edge is labelled with the introduced constraint to generate the child node. Green dotted outlines are used to indicate which node and variable were selected in stages $(1)$ and $(2)$ to lead to stages $(3)$ and $(4)$. The global primal ($P$) and dual ($D$) bounds are increasingly constrained by repeating stages $1$-$4$ until $P$ and $D$ are equal, at which point a provably optimal solution will have been found. Note that for clarity we only show the detailed information needed at each stage, but that this does not indicate any change to the state of the tree.}
    \label{fig:bnb_diagram}
\end{figure*}

\secInLine{Learning-to-branch}
Recent advances in deep learning have led \ac{ml} researchers to contribute to exact \ac{co} (surveys provided by \citealt{Lodi2017}, \citealt{bengio2020machine}, and \citealt{cappart2021combinatorial}). \citealt{khalil2016LearningTB} pioneered the community's interest by using \ac{il} to train a \ac{svm} to imitate the variable rankings of \ac{sb} after the first $500$ \ac{bnb} node visits and thereafter use the \ac{svm}. \citealt{alvarez2017AML} similarly imitated \ac{sb}, but learned to predict the \ac{sb} scores directly using Extremely Randomized Trees \citep{Geurts2006}. These approaches performed promisingly, but their per-instance training and use of \ac{sb} at test time limited their scalability.

These issues were overcome by \citealt{gasse2019}, who 
took as input a bipartite graph representation capturing the current \ac{bnb} node state and predicted the corresponding action chosen by \ac{sb} using a \ac{gcn}. This alleviated the reliance on extensive feature engineering, avoided the use of \ac{sb} at inference time, and demonstrated generalisation to larger instances than seen in training.
Works since have sought to extend this method by introducing new observation features to generalise across heterogeneous \ac{co} instances \citep{zarpellon2021parameterizing} and designing \ac{sb}-on-a-GPU expert labelling methods for scalability \citep{nair2021solving}.


\citealt{etheve2020} proposed \ac{fmsts} which, to the best of our knowledge, is the only published work to apply \ac{rl} to branching and is therefore the \ac{sota} \ac{rl} branching algorithm. By using a \ac{dfs} node selection strategy, they used the \ac{dqn} approach \citep{mnih2013playing} to approximate the Q-function of the \ac{bnb} sub-tree size rooted at the current node; a local Q-function which, in their setting, was equivalent to the number of global tree nodes. Although \ac{fmsts} alleviated issues with credit assignment and partial observability, it relied on using the \ac{dfs} node selection policy (which can be far from optimal), was fundamentally limited by exponential sub-tree sizes produced by larger instances, and its associated models and data sets were not open-accessed.

\section{Background}
\label{sec:background}

\secInLine{Mixed integer linear programming}
An \ac{milp} is an optimisation task where values must be assigned to a set of $n$ \textit{decision variables} subject to a set of $m$ linear \textit{constraints} such that some linear \textit{objective function} is minimised. \Acp{milp} can be written in the standard form

\begin{equation}
    \arg\min_{\bold{x}} \big\{ \bold{c}^{\top} \bold{x} | \bold{A}\bold{x} \leq \bold{b}, \bold{l} \leq \bold{x} \leq \bold{u}, \bold{x} \in \mathbb{Z}^{p} \times \mathbb{R}^{n - p} \big\},
\end{equation}

where $\bold{c} \in \mathbb{R}^{n}$ is a vector of the objective function's coefficients for each decision variable in $\bold{x}$ such that $\bold{c}^{\top}\bold{x}$ is the objective value, $\bold{A} \in \mathbb{R}^{m \times n}$ is a matrix of the $m$ constraints' coefficients (rows) applied to $n$ variables (columns), $\bold{b} \in \mathbb{R}^{m}$ is the vector of variable constraint right-hand side bound values which must be adhered to, and $\bold{l}, \bold{u} \in \mathbb{R}^{n}$ are the respective lower and upper variable value bounds. \Acp{milp} are hard to solve owing to their \textit{integrality constraint(s)} whereby $p \leq n$ decision variables must be an integer. If these integrality constraints are relaxed, the \ac{milp} becomes a \ac{lp}, which can be solved efficiently using algorithms such as simplex \citep{NeldMead65}. The most popular approach for solving \acp{milp} exactly is \ac{bnb}.

\secInLine{Branch-and-bound}
\label{sec:branch_and_bound}
\ac{bnb} is an algorithm composed of multiple heuristics for solving \acp{milp}. It uses a search tree where nodes are \acp{milp} and edges are partition conditions (added constraints) between them. Using a \textit{divide and conquer} strategy, the \ac{milp} is iteratively partitioned into sub-\acp{milp} with smaller solution spaces until an optimal solution (or, if terminated early, a solution with a worst-case optimality gap guarantee) is found. The task of \ac{bnb} is to evolve the search tree until the provably optimal node is found. 

Concretely, as summarised in Figure \ref{fig:bnb_diagram}, at each step in the algorithm, \ac{bnb}: (1) Selects an open (unfathomed leaf) node in the tree whose sub-tree seems promising to evolve; (2) selects (`branches on') a variable to tighten the bounds on the sub-\ac{milp}'s solution space by adding constraints either side of the variable's \ac{lp} solution value, generating two child nodes (sub-\acp{milp}) beneath the focus node; (3) for each child, i) solve the relaxed \ac{lp} (the \textit{dual problem}) to get the \textit{dual bound} (a bound on the best possible objective value in the node's sub-tree) and, where appropriate, ii) solve the \textit{primal problem} and find a feasible (but not necessarily optimal) solution satisfying the node's constraints, thus giving the \textit{primal bound} (the worst-case feasible objective value in the sub-tree); and (4) fathom any children (i.e. consider the sub-tree rooted at the child `fully known' and therefore excluded from any further exploration) whose relaxed \ac{lp} solution is integer-feasible, is worse than the incumbent (the globally best feasible node found so far), or which cannot meet the non-integrality constraints of the \ac{milp}. This process is repeated until the \textit{primal-dual gap} (global primal-dual bound difference) is \num{0}, at which point a provably optimal solution to the original \ac{milp} will have been found. 

Note that the heuristics (i.e.\ primal, branching, and node selection) at each stage jointly determine the performance of \ac{bnb}. More advanced procedures such as cutting planes \citep{Mitchell2009} and column generation \citep{Barnhart1998} are available for enhancement, but are beyond the scope of this work.
Note also that solvers such as \citealt{scip2020} only store `visitable' nodes in memory, therefore in practice fathoming occurs at a feasible node where a branching decision led to the node's two children being outside the established optimality bounds, being infeasible, or having an integer-feasible dual solution, thereby closing the said node's sub-tree.

\secInLine{Q-learning}
Q-learning is typically applied to sequential decision making problems formulated as an \ac{mdp} defined by tuple $\{\mathcal{S}, \mathcal{U}, \mathcal{T}, \mathcal{R}, \gamma \}$. $\mathcal{S}$ is a finite set of states, $\mathcal{U}$ a set of actions, $\mathcal{T}: \mathcal{S} \times \mathcal{U}\times \mathcal{S} \rightarrow [0,1]$ a transition function from state $s \in \mathcal{S}$ to $s^{\prime} \in \mathcal{S}$ given action $u \in {U}$, $\mathcal{R}: \mathcal{S} \rightarrow \mathbb{R}$ a function returning a scalar reward from $s$, and $\gamma \in [0, 1]$ a factor by which to discount expected future returns to their present value. It is an off-policy temporal difference method which aims to learn the action-value function mapping state-action pairs to the expected discounted sum of their immediate and future rewards when following a policy $\pi: \mathcal{S} \rightarrow \mathcal{U}$, $Q^{\pi}(s, u) = \mathbb{E}_{\pi} \big[ \sum_{t^\prime=t+1}^{\infty} \gamma_{t^\prime-1} r(s_{t^\prime}) \vert s_{t} {=} s, u_{t} {=} u \big]$. By definition, an optimal policy $\pi_{*}$ will select an action which maximises the true Q-value $Q_{*}(s, u)$, $\pi_{*}(s) = \arg\max_{u^{\prime}} Q_{*}(s, u^{\prime})$. For scalability, \ac{dqn} \citep{mnih2013playing, vanhasselt2015deep} approximates this true Q-function using a \ac{dnn} parameterised by $\theta$ such that $Q_{\theta}(s, u) \approx Q_{*}(s, u)$.

\section{Retro Branching}
\label{sec:methodology}



We now describe our retro branching approach for learning-to-branch with \ac{rl}.

\secInLine{States}
At each time step $t$ the \ac{bnb} solver state is comprised of the search tree with past branching decisions, per-node \ac{lp} solutions, the global incumbent, the currently focused leaf node, and any other solver statistics which might be tracked.
To convert this information into a suitable input for the branching agent, we represent the \ac{milp} of the focus node chosen by the node selector as a bipartite graph. Concretely, the $n$ variables and $m$ constraints are connected by edges denoting which variables each constraint applies to. This formulation closely follows the approach of \citealt{gasse2019}, with a full list of input features at each node detailed in Appendix \ref{app:observation_features}.

\secInLine{Actions}
Given the \ac{milp} state $s_{t}$ of the current focus node, the branching agent uses a policy $\pi(u_{t} | s_{t})$ to select a variable $u_{t}$ from among the $p$ branching candidates.

\secInLine{Original full episode transitions}
In the original full \ac{bnb} episode, the next node visited is chosen by the node selection policy from amongst any of the open nodes in the tree. This is done independently of the brancher, which observes state information related only to the current focus node and the status of the global bounds. As such, the transitions of the `full episode' are partially observable to the brancher, and it will therefore have the challenging task of needing to aggregate over unobservable states in external sub-trees to predict the long-term values of states and actions.

\secInLine{Retrospectively constructed trajectory transitions (retro branching)}
To address the partial observability of the full episode, we retrospectively construct multiple trajectories where all sequential states in a given trajectory are within the same sub-tree, and where the trajectory's terminal state is chosen from amongst the as yet unchosen fathomed sub-tree leaves. A visualisation of our approach is shown in Figure \ref{fig:subtree_episodes_construction}. Concretely, during training, we first solve the instance as usual with the \ac{rl} brancher and any node selection heuristic to form the `original episode'. When the instance is solved, rather than simply adding the originally observed \ac{mdp}'s transitions to the \ac{dqn} replay buffer, we retrospectively construct multiple trajectory paths through the search tree. This construction process is done by starting at the highest level node not yet added to a trajectory, selecting an as yet unselected fathomed leaf in the sub-tree rooted at said node using some `construction heuristic' (see Section \ref{sec:results_and_discussion}), and using this root-leaf pair as a source-destination with which to construct a path (a `retrospective trajectory'). This process is iteratively repeated until each eligible node in the original search tree has been added to one, and only one, retrospective trajectory. The transitions of each trajectory are then added to the experience replay buffer for learning. Note that retrospective trajectories are only used during training, therefore retro branching agents have no additional inference-time overhead.

Crucially, retro branching determines the sequence of states in each trajectory (i.e. the transition function of the \ac{mdp}) such that the next state(s) observed in a given trajectory will \textit{always} be within the same sub-tree (see Figure \ref{fig:subtree_episodes_construction}) regardless of the node selection policy used in the original \ac{bnb} episode.
Our reasoning behind this idea is that the state(s) beneath the current focus node within its sub-tree will have characteristics (bounds, introduced constraints, etc.) which are strongly related with those of the current node, making them more observable than were the next states to be chosen from elsewhere in the search tree, as can occur in the `original \ac{bnb}' episode. 
Moreover, by correlating the agent's maximum trajectory length with the depth of the tree rather than the total number of nodes, reconstructed trajectories have orders of magnitude fewer steps and lower return variance than the original full episode, making learning tractable on large \acp{milp}.
Furthermore, because the sequential nodes visited are chosen retrospectively in each trajectory, unlike with \ac{fmsts}, any node selection policy can be used during training. As we show in Section \ref{sec:results_and_discussion}, this is a significant help when solving large and complex \acp{milp}.

\secInLine{Rewards}
As demonstrated in Section \ref{sec:results_and_discussion}, the use of reconstructed trajectories enables a simple distance-to-goal reward function to be used; a $r=-1$ punishment is issued to the agent at each step except when the agent's action fathomed the sub-tree, where the agent receives $r=0$. This provides an incentive for the the branching agent to reach the terminal state as quickly as possible. 
When aggregated over all trajectories in a given sub-tree, this auxiliary objective corresponds to fathoming the whole sub-tree (and, by extension, solving the \ac{milp}) in as few steps as possible.
This is because the only nodes which are stored by \citealt{scip2020} and which the brancher will be presented with will be feasible nodes which \textit{potentially} contain the optimal solution beneath them. As such, any action chosen by the brancher which provably shows either the optimal solution to not be beneath the current node or which finds an integer feasible dual solution (i.e. an action which fathoms the sub-tree beneath the node) will be beneficial, because it will prevent SCIP from being able to further needlessly explore the node's sub-tree.

\secInLine{A note on partial observability}
In the above retrospective formulation of the branching \ac{mdp}, the primal, branching, and node selection heuristics active in other sub-trees will still influence the future states and fathoming conditions of a given retrospective trajectory. We posit that there are two extremes; \ac{dfs} node selection where future states are fully observable to the brancher, and non-\ac{dfs} node selection where they are heavily obscured. As shown in Section \ref{sec:results_and_discussion}, our retrospective node selection setting strikes a balance between these two extremes, attaining sufficient observability to facilitate learning while enabling the benefits of short, low variance trajectories with sophisticated node selection strategies which make handling larger \acp{milp} tractable.

\begin{figure*}[tp!]
    \centering
    \includegraphics[width=\textwidth]{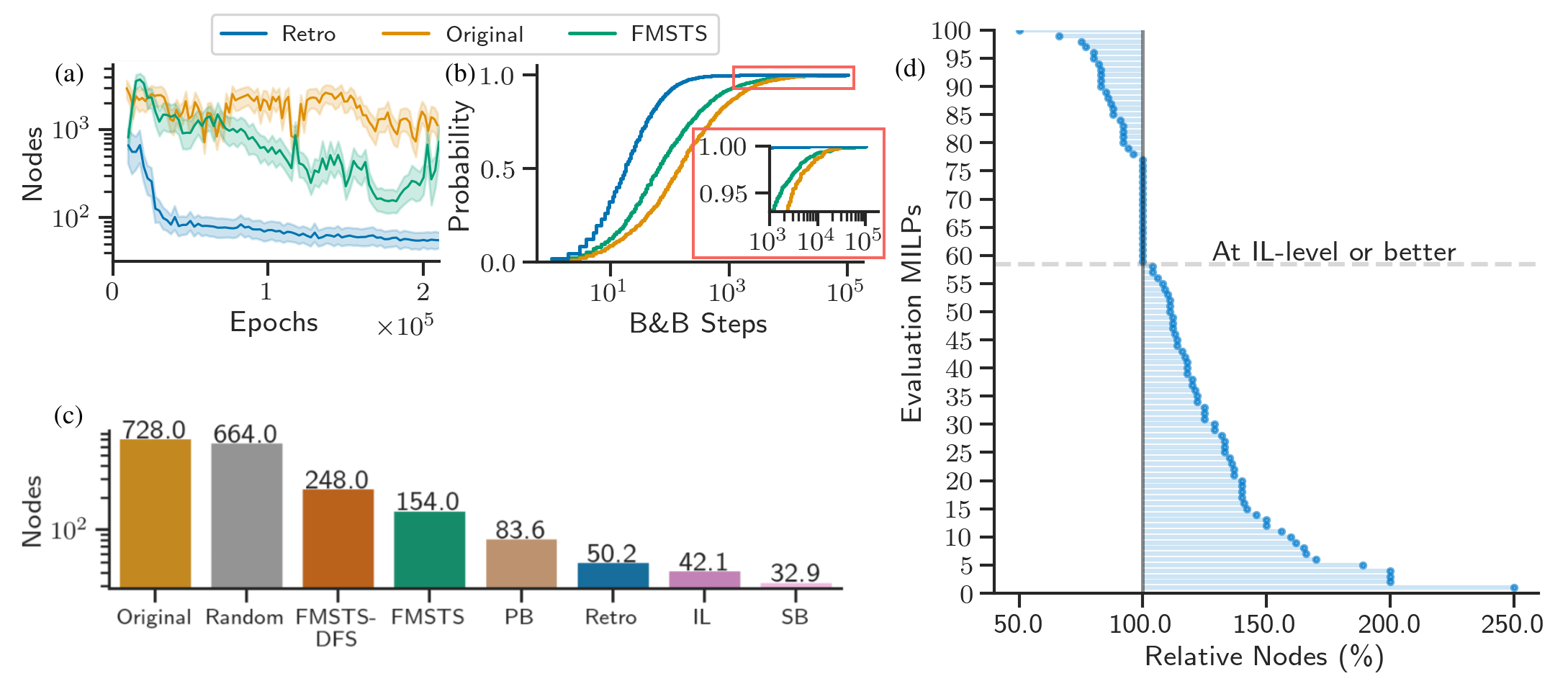}
    \caption{Performances of the branching agents on the $500\times1000$ set covering instances. (a) Validation curves for the \ac{rl} agents evaluated in the same non-\ac{dfs} setting. (b) CDF of the number of \ac{bnb} steps taken by the \ac{rl} agents for each instance seen during training. (c) The best validation performances of each branching agent. (d) The instance-level validation performance of the retro branching agent relative to the \ac{il} agent, with \ac{rl} matching or beating \ac{il} on $42\%$ of test instances.}
    \label{fig:500x1000_branching_performance_plots}
\end{figure*}

\section{Experimental Setup}
\label{sec:experimental_setup}

All code for reproducing the experiments and links to the generated data sets are provided at \url{https://github.com/cwfparsonson/retro_branching}.

\secInLine{Network architecture and learning algorithm}
We used the \ac{gcn} architecture of \citealt{gasse2019} to parameterise the \ac{dqn} value function with some minor modifications which we found to be helpful (see Appendix \ref{app:network_architecture}).
We trained our network with n-step DQN \citep{Sutton1988, mnih2013playing} using prioritised experience replay \citep{schaul2016prioritized}, soft target network updates \citep{lillicrap2019continuous}, and an epsilon-stochastic exploration policy (see Appendix \ref{app:training_parameters} for a detailed description of our \ac{rl} approach and the corresponding algorithms and hyperparameters used).

\secInLine{B\&B environment}
We used the open-source Ecole \citep{prouvost2020ecole} and PySCIPOpt \citep{MaherMiltenbergerPedrosoRehfeldtSchwarzSerrano2016} libraries with SCIP $7.0.1$ \citep{scip2020} as the backend solver to do instance generation and testing. Where possible, we used the training and testing protocols of \citet{gasse2019}. 

\secInLine{MILP Problem classes}
In total, we considered four NP-hard problem benchmarks: set covering \citep{balas_ho_center_2018}, combinatorial auction \citep{LeytonBrown2000}, capacitated facility location \citep{Litvinchev2012}, and maximum independent set \citep{Bergman2016}. 

\secInLine{Baselines}
We compared retro branching against the \ac{sota} \ac{fmsts} \ac{rl} algorithm of \citet{etheve2020} (see Appendix \ref{app:fmsts_implementation} for implementation details) and the \ac{sota} \ac{il} approach of \citet{gasse2019} trained and validated with \num{100000} and \num{20000} strong branching samples respectively. For completeness, we also compared against the \ac{sb} heuristic imitated by the \ac{il} agent, the canonical \ac{pb} heuristic, and a random brancher (equivalent in performance to most infeasible branching \citep{achterberg2005}). Note that we have ommited direct comparison to the \ac{sota} tuned commercial solvers, which we do not claim to be competitive with at this stage. To evaluate the quality of the agents' branching decisions, we used \num{100} validation instances \ins{(see Appendix \ref{app:data_set_size_analysis} for an analysis of this data set size)} which were unseen during training, reporting the total number of tree nodes and \ac{lp} iterations as key metrics to be minimised.

\section{Results \& Discussion}
\label{sec:results_and_discussion}

\subsection{Performance of Retro Branching}

\secInLine{Comparison to the SOTA RL branching heuristics}
We considered set covering instances with \num{500} rows and \num{1000} columns. To demonstrate the benefit of the proposed retro branching method, we trained a baseline `Original' agent on the original full episode, receiving the same reward as our retro branching agent ($-1$ at each non-terminal step and $0$ for a terminal action which ended the episode). We also trained the \ac{sota} \ac{rl} \ac{fmsts} branching agent in a \ac{dfs} setting and, at test time, validated the agent in both a \ac{dfs} (`\ac{fmsts}-\ac{dfs}') and non-\ac{dfs} (`\ac{fmsts}') environment to fairly compare the policies. \ins{Note that the \ac{fmsts} agent serves as an ablation to analyse the influence of training on retrospective trajectories, since it uses our auxiliary objective but without retrospective trajectories, and that the Original agent further ablates the auxiliary objective since its `terminal step' is defined as ending the \ac{bnb} episode (where it receives $r_{t} = 0$ rather than $r_{t}=-1$).} As shown in Figure \ref{fig:500x1000_branching_performance_plots}a, the Original agent was unable to learn on these large instances, with retro branching achieving $14\times$ fewer nodes at test time. \Ac{fmsts} also performed poorly, with highly unstable learning and a final performance $5\times$ and $3\times$ poorer than retro branching in the \ac{dfs} and non-\ac{dfs} settings respectively (see Figure \ref{fig:500x1000_branching_performance_plots}c). We posit that the cause of the poor \ac{fmsts} performance is due to its use of the sub-optimal \ac{dfs} node selection policy, which is ill-suited for handling large \acp{milp} and results in $\approx10\%$ of episodes seen during training being on the order of 10-100\text{k} steps long (see Figure \ref{fig:500x1000_branching_performance_plots}b), which makes learning significantly harder for \ac{rl}.

\secInLine{Comparison to non-RL branching heuristics}
Having demonstrated that the proposed retro branching method makes learning-to-branch at scale tractable for \ac{rl}, we now compare retro branching with the baseline branchers to understand the efficacy of \ac{rl} in the context of the current literature. Figure \ref{fig:500x1000_branching_performance_plots}c shows how retro branching compares to other policies on large $500\times1000$ set covering instances. While the agent outperforms \ac{pb}, it only matches or beats \ac{il} on $42\%$ of the test instances (see Figure \ref{fig:500x1000_branching_performance_plots}d) and, on average, has a $\approx20\%$ larger \ac{bnb} tree size. Therefore although our \ac{rl} agent was still improving and was limited by compute (see Appendix \ref{app:training_time_and_convergence}), and in spite of our method outperforming the current \ac{sota} \ac{fmsts} \ac{rl} brancher, \ac{rl} has not yet been able to match or surpass the \ac{sota} \ac{il} agent at scale. This will be an interesting area of future work, as discussed in Section \ref{sec:conclusion}.

\subsection{Analysis of Retro Branching}

\secInLine{Verifying that RL can outperform IL}
In addition to not needing labelled data, a key motivation for using \ac{rl} over \ac{il} for learning-to-branch is the potential to discover superior policies. While Figure \ref{fig:500x1000_branching_performance_plots} showed that, at test-time, retro branching matched or outperformed \ac{il} on $42\%$ of instances, \ac{il} still had a lower average tree size. As shown in Table \ref{tab:branching_method_evaluation}, we found that, on small set covering instances with \num{165} constraints and \num{230} variables, \ac{rl} could outperform \ac{il} by $\approx20\%$. While improvement on problems of this scale is not the primary challenge facing \ac{ml}-\ac{bnb} solvers, we are encouraged by this demonstration that it is possible for an \ac{rl} agent to learn a policy better able to maximise the performance of an expressivity-constrained network than imitating an expert such as \ac{sb} without the need for pre-training or expensive data labelling procedures (see Appendix \ref{app:cost_strong_branching_labels}). 

    
        


For completeness, Table \ref{tab:branching_method_evaluation} also compares the retro branching agent to the \ac{il}, \ac{pb}, and \ac{sb} branching policies evaluated on \num{100} unseen instances of four NP-hard \ac{co} benchmarks. We considered instances with \num{10} items and \num{50} bids for combinatorial auction, \num{5} customers and facilities for capacitated facility location, and \num{25} nodes for maximum independent set. \Ac{rl} achieved a lower number of tree nodes than \ac{pb} and \ac{il} on all problems except combinatorial auction. This highlights the potential for \ac{rl} to learn improved branching policies to solve a variety of \acp{milp}.

\begin{table*}[!tp]
	\caption{
	Test-time comparison of the best agents on the evaluation instances of the four NP-hard small \ac{co} problems considered.
	}
	\small
	\centering
	\resizebox{\textwidth}{!}{%
	\begin{tabular}{c|cc|cc|cc|cc}
	
	\toprule
	
	\multicolumn{1}{c}{} &
	\multicolumn{2}{c}{Set Covering} &
	\multicolumn{2}{c}{Combinatorial Auction} &
	\multicolumn{2}{c}{Capacitated Facility Location} &
	\multicolumn{2}{c}{Maximum Independent Set}
	\\
	
	\cmidrule[0.75pt](lr){2-3}
	\cmidrule[0.75pt](lr){4-5}
	\cmidrule[0.75pt](lr){6-7}
	\cmidrule[0.75pt](lr){8-9}
    
    Method 
    & \# LPs & \# Nodes
    & \# LPs & \# Nodes
    & \# LPs & \# Nodes
    & \# LPs & \# Nodes
	\\
	
	\cmidrule[0.75pt](lr){1-9}
	
	\ac{sb}
	& $184$ & $6.76$
	& $13.2$ & $4.64$
	& $28.2$ & $10.2$
	& $19.2$ & $3.80$
	\\
	
	\cmidrule[0.05pt](lr){1-9}
	
	
	\ac{pb} 
	& $258$ & $12.8$
	& $22.0$ & $7.80$
	& $\mathbf{28.0}$ & $10.2$
	& $25.4$ & $5.77$
	\\
	
	\ac{il} 
	& $244$ & $10.5$
	& $\mathbf{16.0}$ & $\mathbf{5.29}$
	& $\mathbf{28.0}$ & $10.2$
	& $20.1$ & $4.08$
	\\
	
	Retro
	& $\mathbf{206}$ & $\mathbf{8.68}$
	& $18.1$ & $5.73$
	& $28.4$ & $\mathbf{10.1}$
	& $\mathbf{19.1}$ & $\mathbf{4.01}$
	\\

	\bottomrule
	
	\end{tabular}
	} 
	\label{tab:branching_method_evaluation}
\end{table*}

\begin{figure*}[tp!]
    \centering
    \includegraphics[width=\textwidth]{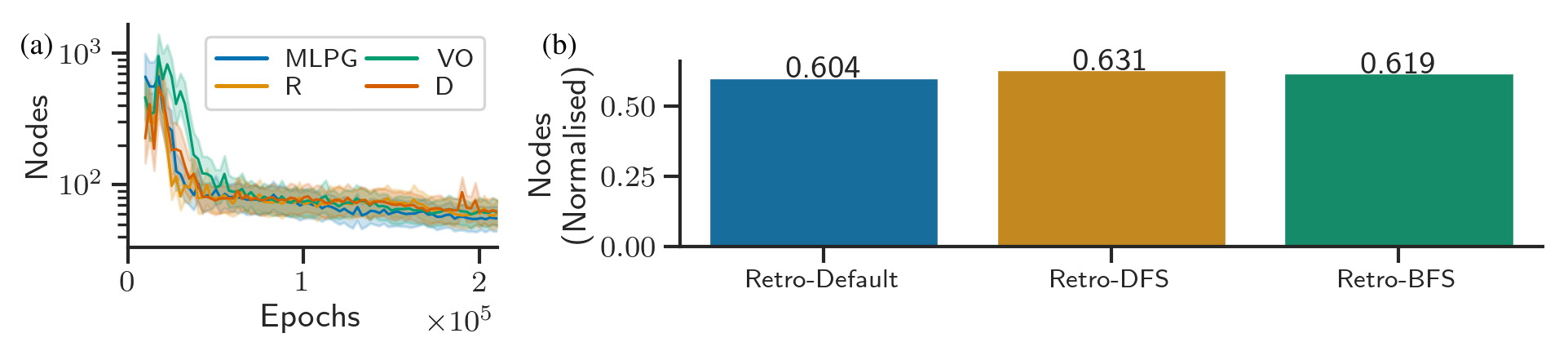}
    \caption{$500\times1000$ set covering performances. (a) Validation curves for four retro branching agents each trained with a different trajectory construction heuristic: Maximum LP gain (MLPG); random (R); visitation order (VO); and deepest (D). (b) The performances of the best retro branching agent deployed in three different node selection environments (default SCIP, \ac{dfs}, and \ac{bfs}) normalised relative to the performances of \ac{pb} (measured by number of tree nodes).}
    \label{fig:normalised_rl_performance_in_different_node_selection_settings}
\end{figure*}

\secInLine{Demonstrating the independence of retro branching to future state selection}
As described in Section \ref{sec:methodology}, in order to retrospectively construct a path through the search tree, a fathomed leaf node must be selected. We refer to the method for selecting the leaf node as the \textit{construction heuristic}. The future states seen by the agent are therefore determined by the construction heurisitc (used in training) and the node selection heuristic (used in training and inference). 

During our experiments, we found that the specific construction heuristic used had little impact on the performance of our agent. Figure \ref{fig:normalised_rl_performance_in_different_node_selection_settings}a shows the validation curves for four agents trained on $500\times1000$ set covering instances each using one of the following construction heuristics: Maximum LP gain (`MLPG': Select the leaf with the largest LP gain); random (`R': Randomly select a leaf); visitation order (`VO': Select the leaf which was visited first in the original episode); and deepest (`D': Select the leaf which results in the longest trajectory). As shown, all construction heuristics resulted in roughly the same performance (with MLPG performing only slightly better). 
This suggests that the agent learns to reduce the trajectory length regardless of the path chosen by the construction heuristic. Since the specific path chosen is independent of node selection, we posit that the relative strength of an \ac{rl} agent trained with retro branching will also be independent of the node selection policy used.

To test this, we took our best retro branching agent trained with the default SCIP node selection heuristic and tested it on the $500\times1000$ validation instances in the default, \ac{dfs}, and \ac{bfs} SCIP node selection settings. To make the performances of the brancher comparable across these settings, we normalised the mean tree sizes with those of \ac{pb} (a branching heuristic independent of the node selector) to get the performance relative to \ac{pb} in each environment. As shown in Figure \ref{fig:normalised_rl_performance_in_different_node_selection_settings}b, our agent achieved consistent relative performance regardless of the node selection policy used, indicating its indifference to the node selector.




\section{Conclusions}
\label{sec:conclusion}

We have introduced retro branching; a retrospective approach to constructing \ac{bnb} trajectories in order to aid learning-to-branch with \ac{rl}. We posited that retrospective trajectories address the challenges of long episodes, large state-action spaces, and partially observable future states which otherwise make branching an acutely difficult task for \ac{rl}. We empirically demonstrated that retro branching outperforms the current \ac{sota} \ac{rl} method by $3$-$5\times$ and comes within $20\%$ of the performance of \ac{il} whilst matching or beating it on $42\%$ of test instances. Moreover, we showed that \ac{rl} can surpass the performance of \ac{il} on small instances, exemplifying a key advantage of \ac{rl} in being able to discover novel performance-maximising policies for expressivity-constrained networks without the need for pre-training or expert examples. However, retro branching was not able to exceed the \ac{il} agent at scale. In this section we outline areas of further work.

\secInLine{Partial observability} 
A limitation of our proposed approach is the remaining partial observability of the \ac{mdp}, with activity external to the current sub-tree and branching decision influencing future bounds, states, and rewards. In this and other studies, variable and node selection have been considered in isolation. An interesting approach would be to combine node and variable selection, giving the agent full control over how the \ac{bnb} tree is evolved.


\secInLine{Reward function}
The proposed trajectory reconstruction approach can facilitate a simple \ac{rl} reward function which would otherwise fail were the original `full' tree episode used. However, assigning a $-1$ reward at each step in a given trajectory ignores the fact that certain actions, particularly early on in the \ac{bnb} process, can have significant influence over the length of multiple trajectories. This could be accounted for in the reward signal, perhaps by using a retrospective backpropagation method (similar to value backpropagation in Monte Carlo tree search \citep{Silver2016, Silver2017}).

\secInLine{Exploration}
The large state-action space and the complexity of making thousands of sequential decisions which together influence final performance in complex ways makes exploration in \ac{bnb} an acute challenge for \ac{rl}. One reason for \ac{rl} struggling to close the $20\%$ performance gap with \ac{il} at scale could be that, at some point, stochastic action sampling to explore new policies is highly unlikely to find trajectories with improved performance. As such, more sophisticated exploration strategies could be promising, such as novel experience intrinsic reward signals \citep{burda2018exploration, zhang2021noveld}, reverse backtracking through the episode to improve trajectory quality \citep{salimans2018learning, Agostinelli2019, ecoffet2021goexplore}, and avoiding local optima using auxiliary distance-to-goal rewards \citep{Trott2019KeepingYD} or evolutionary strategies \citep{Conti2018}.



\section{Acknowledgments}

We would like to thank Maxime Gasse, Antoine Prouvost, and the rest of the Ecole development team for answering our questions on SCIP and Ecole, and also the anonymous reviewers for their constructive comments on earlier versions of this paper.

\bibliography{aaai23}

\begin{thebibliography}{52}
\providecommand{\natexlab}[1]{#1}

\bibitem[{Achterberg(2007)}]{achterberg2007}
Achterberg, T. 2007.
\newblock \emph{{Constraint Integer Programming}}.
\newblock Doctoral thesis, Technische Universit{\"{a}}t Berlin, Fakult{\"{a}}t
  II - Mathematik und Naturwissenschaften, Berlin.

\bibitem[{Achterberg, Koch, and Martin(2004)}]{achterberg2005}
Achterberg, T.; Koch, T.; and Martin, A. 2004.
\newblock {Branching rules revisited}.
\newblock Technical Report 04-13, ZIB, Takustr. 7, 14195 Berlin.

\bibitem[{Achterberg and Wunderling(2013)}]{Achterberg2013}
Achterberg, T.; and Wunderling, R. 2013.
\newblock \emph{{Mixed Integer Programming: Analyzing 12 Years of Progress}}.

\bibitem[{Agostinelli et~al.(2019)Agostinelli, McAleer, Shmakov, and
  Baldi}]{Agostinelli2019}
Agostinelli, F.; McAleer, S.; Shmakov, A.; and Baldi, P. 2019.
\newblock Solving the Rubik's cube with deep reinforcement learning and search.
\newblock \emph{Nature Machine Intelligence}, 1(8): 356--363.

\bibitem[{Alvarez, Louveaux, and Wehenkel(2017)}]{alvarez2017AML}
Alvarez, A.~M.; Louveaux, Q.; and Wehenkel, L. 2017.
\newblock A Machine Learning-Based Approximation of Strong Branching.
\newblock \emph{INFORMS J. Comput.}, 29: 185--195.

\bibitem[{Applegate et~al.(2007)Applegate, Bixby, Chvatal, and
  Cook}]{Applegate2007}
Applegate, D.~L.; Bixby, R.~E.; Chvatal, V.; and Cook, W.~J. 2007.
\newblock \emph{{The Traveling Salesman Problem: A Computational Study}}.
\newblock Princeton University Press.

\bibitem[{Applegate et~al.(1995)Applegate, Bixpy, Chvatal, and
  Cook}]{Applegate1995}
Applegate, D.~L.; Bixpy, R.~E.; Chvatal, V.; and Cook, W.~J. 1995.
\newblock {Finding cuts in the TSP (A preliminary report)}.
\newblock Technical report, DIMACS.

\bibitem[{Balas, Ho, and Center(2018)}]{balas_ho_center_2018}
Balas, E.; Ho, A.; and Center, C. M. U.~R. 2018.
\newblock Set covering algorithms using cutting planes, heuristics, and
  subgradient optimization : a computational study.

\bibitem[{Barahona(1982)}]{barahona82}
Barahona, F. 1982.
\newblock On the computational complexity of {Ising} spin glass models.
\newblock \emph{Journal of Physics A: Mathematical and General}, 15(10): 3241.

\bibitem[{Barnhart et~al.(1998)Barnhart, Johnson, Nemhauser, Savelsbergh, and
  Vance}]{Barnhart1998}
Barnhart, C.; Johnson, E.~L.; Nemhauser, G.~L.; Savelsbergh, M. W.~P.; and
  Vance, P.~H. 1998.
\newblock Branch-And-Price: Column Generation for Solving Huge Integer
  Programs.
\newblock \emph{Oper. Res.}, 46(3): 316–329.

\bibitem[{Bengio, Lodi, and Prouvost(2021)}]{bengio2020machine}
Bengio, Y.; Lodi, A.; and Prouvost, A. 2021.
\newblock {Machine learning for combinatorial optimization: A methodological
  tour d'horizon}.
\newblock \emph{European Journal of Operational Research}, 290(2): 405--421.

\bibitem[{Benichou et~al.(1971)Benichou, Gauthier, Girodet, Hentges, Ribiere,
  and Vincent}]{Benichou1971}
Benichou, M.; Gauthier, J.~M.; Girodet, P.; Hentges, G.; Ribiere, G.; and
  Vincent, O. 1971.
\newblock {Experiments in mixed-integer linear programming}.
\newblock \emph{Mathematical Programming}, 1(1): 76--94.

\bibitem[{Bergman et~al.(2016)Bergman, Cire, Hoeve, and Hooker}]{Bergman2016}
Bergman, D.; Cire, A.~A.; Hoeve, W.-J.~v.; and Hooker, J. 2016.
\newblock \emph{Decision Diagrams for Optimization}.
\newblock Springer Publishing Company, Incorporated, 1st edition.
\newblock ISBN 3319428470.

\bibitem[{Burda et~al.(2018)Burda, Edwards, Storkey, and
  Klimov}]{burda2018exploration}
Burda, Y.; Edwards, H.; Storkey, A.; and Klimov, O. 2018.
\newblock Exploration by Random Network Distillation.

\bibitem[{Cappart et~al.(2021)Cappart, Chételat, Khalil, Lodi, Morris, and
  Veličković}]{cappart2021combinatorial}
Cappart, Q.; Chételat, D.; Khalil, E.; Lodi, A.; Morris, C.; and Veličković,
  P. 2021.
\newblock Combinatorial optimization and reasoning with graph neural networks.

\bibitem[{Conti et~al.(2018)Conti, Madhavan, Such, Lehman, Stanley, and
  Clune}]{Conti2018}
Conti, E.; Madhavan, V.; Such, F.~P.; Lehman, J.; Stanley, K.~O.; and Clune, J.
  2018.
\newblock Improving Exploration in Evolution Strategies for Deep Reinforcement
  Learning via a Population of Novelty-Seeking Agents.
\newblock In \emph{Proceedings of the 32nd International Conference on Neural
  Information Processing Systems}, NIPS'18, 5032–5043. Red Hook, NY, USA:
  Curran Associates Inc.

\bibitem[{CPLEX(2009)}]{cplex2009v12}
CPLEX, I. 2009.
\newblock V12. 1: User’s Manual for CPLEX.
\newblock \emph{International Business Machines Corporation}, 46(53): 157.

\bibitem[{Ecoffet et~al.(2021)Ecoffet, Huizinga, Lehman, Stanley, and
  Clune}]{ecoffet2021goexplore}
Ecoffet, A.; Huizinga, J.; Lehman, J.; Stanley, K.~O.; and Clune, J. 2021.
\newblock First return, then explore.
\newblock \emph{Nature}, 590(7847): 580–586.

\bibitem[{Etheve et~al.(2020)Etheve, Alès, Bissuel, Juan, and
  Kedad-Sidhoum}]{etheve2020}
Etheve, M.; Alès, Z.; Bissuel, C.; Juan, O.; and Kedad-Sidhoum, S. 2020.
\newblock Reinforcement Learning for Variable Selection in a Branch and Bound
  Algorithm.
\newblock \emph{Lecture Notes in Computer Science}, 176–185.

\bibitem[{Gasse et~al.(2019)Gasse, Ch\'{e}telat, Ferroni, Charlin, and
  Lodi}]{gasse2019}
Gasse, M.; Ch\'{e}telat, D.; Ferroni, N.; Charlin, L.; and Lodi, A. 2019.
\newblock \emph{Exact Combinatorial Optimization with Graph Convolutional
  Neural Networks}.
\newblock Red Hook, NY, USA: Curran Associates Inc.

\bibitem[{Geurts, Ernst, and Wehenkel(2006)}]{Geurts2006}
Geurts, P.; Ernst, D.; and Wehenkel, L. 2006.
\newblock {Extremely randomized trees}.
\newblock \emph{Machine Learning}, 63(1): 3--42.

\bibitem[{Harutyunyan et~al.(2019)Harutyunyan, Dabney, Mesnard, Heess, Azar,
  Piot, van Hasselt, Singh, Wayne, Precup, and Munos}]{Harutyunyan2019}
Harutyunyan, A.; Dabney, W.; Mesnard, T.; Heess, N.; Azar, M.~G.; Piot, B.; van
  Hasselt, H.; Singh, S.; Wayne, G.; Precup, D.; and Munos, R. 2019.
\newblock \emph{Hindsight Credit Assignment}.
\newblock Red Hook, NY, USA: Curran Associates Inc.

\bibitem[{He, Daum\'{e}, and Eisner(2014)}]{He2014}
He, H.; Daum\'{e}, H.; and Eisner, J. 2014.
\newblock Learning to Search in Branch-and-Bound Algorithms.
\newblock In \emph{Proceedings of the 27th International Conference on Neural
  Information Processing Systems - Volume 2}, NIPS'14, 3293–3301. Cambridge,
  MA, USA: MIT Press.

\bibitem[{Hessel et~al.(2017)Hessel, Modayil, van Hasselt, Schaul, Ostrovski,
  Dabney, Horgan, Piot, Azar, and Silver}]{hessel2017rainbow}
Hessel, M.; Modayil, J.; van Hasselt, H.; Schaul, T.; Ostrovski, G.; Dabney,
  W.; Horgan, D.; Piot, B.; Azar, M.; and Silver, D. 2017.
\newblock Rainbow: Combining Improvements in Deep Reinforcement Learning.

\bibitem[{Khalil et~al.(2016)Khalil, Bodic, Song, Nemhauser, and
  Dilkina}]{khalil2016LearningTB}
Khalil, E.~B.; Bodic, P.~L.; Song, L.; Nemhauser, G.~L.; and Dilkina, B.~N.
  2016.
\newblock Learning to Branch in Mixed Integer Programming.
\newblock In \emph{AAAI}.

\bibitem[{Korte and Vygen(2012)}]{korte12}
Korte, B.~H.; and Vygen, J. 2012.
\newblock \emph{Combinatorial Optimization: Theory and Algorithms}.
\newblock New York, NY: Springer-Verlag.
\newblock ISBN 9783642244889 3642244882 3642244874 9783642244872.

\bibitem[{Land and Doig(1960)}]{land1960}
Land, A.~H.; and Doig, A.~G. 1960.
\newblock An Automatic Method of Solving Discrete Programming Problems.
\newblock \emph{Econometrica}, 28(3): pp. 497--520.

\bibitem[{Leyton-Brown, Pearson, and Shoham(2000)}]{LeytonBrown2000}
Leyton-Brown, K.; Pearson, M.; and Shoham, Y. 2000.
\newblock Towards a Universal Test Suite for Combinatorial Auction Algorithms.
\newblock In \emph{Proceedings of the 2nd ACM Conference on Electronic
  Commerce}, EC '00, 66–76. New York, NY, USA: Association for Computing
  Machinery.
\newblock ISBN 1581132727.

\bibitem[{Lillicrap et~al.(2019)Lillicrap, Hunt, Pritzel, Heess, Erez, Tassa,
  Silver, and Wierstra}]{lillicrap2019continuous}
Lillicrap, T.~P.; Hunt, J.~J.; Pritzel, A.; Heess, N.; Erez, T.; Tassa, Y.;
  Silver, D.; and Wierstra, D. 2019.
\newblock Continuous control with deep reinforcement learning.

\bibitem[{Lin(1992)}]{Lin1992}
Lin, L.-J. 1992.
\newblock Self-Improving Reactive Agents Based on Reinforcement Learning,
  Planning and Teaching.
\newblock \emph{Mach. Learn.}, 8(3–4): 293–321.

\bibitem[{Litvinchev and Ozuna~Espinosa(2012)}]{Litvinchev2012}
Litvinchev, I.; and Ozuna~Espinosa, E.~L. 2012.
\newblock Solving the Two-Stage Capacitated Facility Location Problem by the
  Lagrangian Heuristic.
\newblock In Hu, H.; Shi, X.; Stahlbock, R.; and Vo{\ss}, S., eds.,
  \emph{Computational Logistics}, 92--103. Berlin, Heidelberg: Springer Berlin
  Heidelberg.
\newblock ISBN 978-3-642-33587-7.

\bibitem[{Lodi and Zarpellon(2017)}]{Lodi2017}
Lodi, A.; and Zarpellon, G. 2017.
\newblock {On learning and branching: a survey}.
\newblock \emph{TOP}, 25(2): 207--236.

\bibitem[{Maher et~al.(2016)Maher, Miltenberger, Pedroso, Rehfeldt, Schwarz,
  and Serrano}]{MaherMiltenbergerPedrosoRehfeldtSchwarzSerrano2016}
Maher, S.; Miltenberger, M.; Pedroso, J.~P.; Rehfeldt, D.; Schwarz, R.; and
  Serrano, F. 2016.
\newblock {PySCIPOpt}: Mathematical Programming in Python with the {SCIP}
  Optimization Suite.
\newblock In \emph{Mathematical Software {\textendash} {ICMS} 2016}, 301--307.
  Springer International Publishing.

\bibitem[{Mao et~al.(2019)Mao, Venkatakrishnan, Schwarzkopf, and
  Alizadeh}]{Mao2019}
Mao, H.; Venkatakrishnan, S.~B.; Schwarzkopf, M.; and Alizadeh, M. 2019.
\newblock Variance Reduction for Reinforcement Learning in Input-Driven
  Environments.
\newblock In \emph{7th International Conference on Learning Representations,
  {ICLR} 2019, New Orleans, LA, USA, May 6-9, 2019}. OpenReview.net.

\bibitem[{Mitchell(2009)}]{Mitchell2009}
Mitchell, J.~E. 2009.
\newblock \emph{{Integer programming: branch and cut algorithmsInteger
  Programming: Branch and Cut Algorithms}}, 1643--1650.
\newblock Boston, MA: Springer US.
\newblock ISBN 978-0-387-74759-0.

\bibitem[{Mnih et~al.(2013)Mnih, Kavukcuoglu, Silver, Graves, Antonoglou,
  Wierstra, and Riedmiller}]{mnih2013playing}
Mnih, V.; Kavukcuoglu, K.; Silver, D.; Graves, A.; Antonoglou, I.; Wierstra,
  D.; and Riedmiller, M. 2013.
\newblock Playing Atari with Deep Reinforcement Learning.

\bibitem[{Nair et~al.(2021)Nair, Bartunov, Gimeno, von Glehn, Lichocki, Lobov,
  O'Donoghue, Sonnerat, Tjandraatmadja, Wang, Addanki, Hapuarachchi, Keck,
  Keeling, Kohli, Ktena, Li, Vinyals, and Zwols}]{nair2021solving}
Nair, V.; Bartunov, S.; Gimeno, F.; von Glehn, I.; Lichocki, P.; Lobov, I.;
  O'Donoghue, B.; Sonnerat, N.; Tjandraatmadja, C.; Wang, P.; Addanki, R.;
  Hapuarachchi, T.; Keck, T.; Keeling, J.; Kohli, P.; Ktena, I.; Li, Y.;
  Vinyals, O.; and Zwols, Y. 2021.
\newblock Solving Mixed Integer Programs Using Neural Networks.

\bibitem[{Nelder and Mead(1965)}]{NeldMead65}
Nelder, J.~A.; and Mead, R. 1965.
\newblock A simplex method for function minimization.
\newblock \emph{Computer Journal}, 7: 308--313.

\bibitem[{Perdomo-Ortiz et~al.(2012)Perdomo-Ortiz, Dickson, Drew-Brook, Rose,
  and Aspuru-Guzik}]{perdomo12}
Perdomo-Ortiz, A.; Dickson, N.; Drew-Brook, M.; Rose, G.; and Aspuru-Guzik, A.
  2012.
\newblock Finding low-energy conformations of lattice protein models by quantum
  annealing.
\newblock \emph{Scientific Reports}, 2: 571.

\bibitem[{Prouvost et~al.(2020)Prouvost, Dumouchelle, Scavuzzo, Gasse,
  Ch{\'e}telat, and Lodi}]{prouvost2020ecole}
Prouvost, A.; Dumouchelle, J.; Scavuzzo, L.; Gasse, M.; Ch{\'e}telat, D.; and
  Lodi, A. 2020.
\newblock Ecole: A Gym-like Library for Machine Learning in Combinatorial
  Optimization Solvers.
\newblock In \emph{Learning Meets Combinatorial Algorithms at NeurIPS2020}.

\bibitem[{Salimans and Chen(2018)}]{salimans2018learning}
Salimans, T.; and Chen, R. 2018.
\newblock Learning Montezuma's Revenge from a Single Demonstration.

\bibitem[{Schaul et~al.(2016)Schaul, Quan, Antonoglou, and
  Silver}]{schaul2016prioritized}
Schaul, T.; Quan, J.; Antonoglou, I.; and Silver, D. 2016.
\newblock Prioritized Experience Replay.

\bibitem[{SCIP(2022)}]{scip2020}
SCIP. 2022.
\newblock {The SCIP Optimization Suite 7.0}.
\newblock Technical report, Optimization Online.

\bibitem[{Silver et~al.(2016)Silver, Huang, Maddison, Guez, Sifre, van~den
  Driessche, Schrittwieser, Antonoglou, Panneershelvam, Lanctot, Dieleman,
  Grewe, Nham, Kalchbrenner, Sutskever, Lillicrap, Leach, Kavukcuoglu, Graepel,
  and Hassabis}]{Silver2016}
Silver, D.; Huang, A.; Maddison, C.~J.; Guez, A.; Sifre, L.; van~den Driessche,
  G.; Schrittwieser, J.; Antonoglou, I.; Panneershelvam, V.; Lanctot, M.;
  Dieleman, S.; Grewe, D.; Nham, J.; Kalchbrenner, N.; Sutskever, I.;
  Lillicrap, T.; Leach, M.; Kavukcuoglu, K.; Graepel, T.; and Hassabis, D.
  2016.
\newblock Mastering the game of Go with deep neural networks and tree search.
\newblock \emph{Nature}, 529(7587): 484--489.

\bibitem[{Silver et~al.(2017)Silver, Schrittwieser, Simonyan, Antonoglou,
  Huang, Guez, Hubert, Baker, Lai, Bolton, Chen, Lillicrap, Hui, Sifre, van~den
  Driessche, Graepel, and Hassabis}]{Silver2017}
Silver, D.; Schrittwieser, J.; Simonyan, K.; Antonoglou, I.; Huang, A.; Guez,
  A.; Hubert, T.; Baker, L.; Lai, M.; Bolton, A.; Chen, Y.; Lillicrap, T.; Hui,
  F.; Sifre, L.; van~den Driessche, G.; Graepel, T.; and Hassabis, D. 2017.
\newblock Mastering the game of Go without human knowledge.
\newblock \emph{Nature}, 550(7676): 354--359.

\bibitem[{Sutton(1988)}]{Sutton1988}
Sutton, R.~S. 1988.
\newblock {Learning to predict by the methods of temporal differences}.
\newblock \emph{Machine Learning}, 3(1): 9--44.

\bibitem[{Sutton and Barto(2018)}]{Sutton1998}
Sutton, R.~S.; and Barto, A.~G. 2018.
\newblock \emph{Reinforcement Learning: An Introduction}.
\newblock The MIT Press, second edition.

\bibitem[{Tomazella and Nagano(2020)}]{Tomazella2020}
Tomazella, C.~P.; and Nagano, M.~S. 2020.
\newblock A comprehensive review of Branch-and-Bound algorithms: Guidelines and
  directions for further research on the flowshop scheduling problem.
\newblock \emph{Expert Systems with Applications}, 158: 113556.

\bibitem[{Trott et~al.(2019)Trott, Zheng, Xiong, and
  Socher}]{Trott2019KeepingYD}
Trott, A.; Zheng, S.; Xiong, C.; and Socher, R. 2019.
\newblock Keeping Your Distance: Solving Sparse Reward Tasks Using
  Self-Balancing Shaped Rewards.
\newblock In \emph{NeurIPS}.

\bibitem[{van Hasselt, Guez, and Silver(2015)}]{vanhasselt2015deep}
van Hasselt, H.; Guez, A.; and Silver, D. 2015.
\newblock Deep Reinforcement Learning with Double Q-learning.

\bibitem[{Zarpellon et~al.(2021)Zarpellon, Jo, Lodi, and
  Bengio}]{zarpellon2021parameterizing}
Zarpellon, G.; Jo, J.; Lodi, A.; and Bengio, Y. 2021.
\newblock Parameterizing Branch-and-Bound Search Trees to Learn Branching
  Policies.
\newblock arXiv:2002.05120.

\bibitem[{Zhang et~al.(2021)Zhang, Xu, Wang, Wu, Keutzer, Gonzalez, and
  Tian}]{zhang2021noveld}
Zhang, T.; Xu, H.; Wang, X.; Wu, Y.; Keutzer, K.; Gonzalez, J.~E.; and Tian, Y.
  2021.
\newblock NovelD: A Simple yet Effective Exploration Criterion.
\newblock In Beygelzimer, A.; Dauphin, Y.; Liang, P.; and Vaughan, J.~W., eds.,
  \emph{Advances in Neural Information Processing Systems}.

\end{thebibliography}

\appendix


\section{RL Training}

\subsection{Training Parameters}
\label{app:training_parameters}

The \ac{rl} training hyperparameters are summarised in Table \ref{tab:training_parameters}. We used n-step \ac{dqn} \citep{Sutton1988, mnih2013playing} with prioritised experience replay \citep{schaul2016prioritized}, with overviews of each of these approaches provided below. For exploration, we followed an $\epsilon$-stochastic policy ($\epsilon \in [0, 1]$) whereby the probabilities for action selection were $\epsilon$ for a random action and $1-\epsilon$ for an action sampled from the softmax probability distribution over the Q-values of the branching candidates. We also found it helpful for learning stability to clip the gradients of our network before applying parameter updates.

\begin{table}[!htp]
    \centering
    \begin{tabular}{lr}
    \toprule
         Training Parameter & Value \\
         
         \midrule
         
         Batch size & \num{64} (\num{128}) \\
         Actor steps per learner update & \num{5} (\num{10}) \\
         Learning rate & \num{5e-5} \\
         Discount factor & \num{0.99} \\
         Optimiser & Adam \\
         Buffer size $|\mathcal{M}|_{\text{init}}$ & \num{20e3} \\
         Buffer size $|\mathcal{M}|_{\text{capacity}}$ & \num{100e3} \\
         Prioritised experience replay $\beta_{\text{init}}$ & \num{0.4} \\
         Prioritised experience replay $\beta_{\text{final}}$ & \num{1.0} \\
         $\beta_{\text{init}} \xrightarrow{} \beta_{\text{final}}$ learner steps & \num{5e3} \\
         Prioritised experience replay $\alpha$ & \num{0.6} \\
         Minimum experience priority & \num{1e-3} \\
         Soft target network update $\tau_{\text{soft}}$ & \num{1e-4} \\
         Gradient clip value & \num{10} \\
         n-step \ac{dqn} $n$ & \num{3} \\
         Exploration probability $\epsilon$ & \num{2.5e-2}

    \end{tabular}
    \caption{Training parameters used for training the \ac{rl} agent. All parameters were kept the same across \ac{co} instances except for the large $500 \times 1000$ set covering instances, which we used a larger batch size and actor steps per learner update (specified in brackets).}
    \label{tab:training_parameters}
\end{table}

\secInLine{Conventional \ac{dqn}}
At each time step $t$ during training, $Q_{\theta}(s, u)$ is used with an exploration strategy to select an action and add the observed transition $T = (s_{t}, u_{t}, r_{t+1}, \gamma_{t+1}, s_{t+1})$ to a replay memory buffer \citep{Lin1992}. The network's parameters $\theta$ are then optimised with stochastic gradient descent to minimise the mean squared error loss between the \textit{online} network's predictions and a bootstrapped estimate of the Q-value,

\begin{equation}
    J_{DQN}(Q) = \big[ r_{t+1} + \gamma_{t+1} \max_{u^{\prime}} Q_{\bar{\theta}}(s_{t+1}, u^{\prime}) - Q_{\theta}(s_{t}, u_{t}) \big]^{2},
\end{equation}

where $t$ is a time step randomly sampled from the buffer and $Q_{\bar{\theta}}$ a \textit{target} network with parameters $\bar{\theta}$ which are periodically copied from the acting online network. The target network is not directly optimised, but is used to provide the bootstrapped Q-value estimates for the loss function.

\secInLine{Prioritised experience replay}
Vanilla DQN replay buffers are sampled uniformly to obtain transitions for network updates. A preferable approach is to more frequently sample transitions from which there is much to learn. Prioritised experience replay \citep{schaul2016prioritized} deploys this intuition by sampling transitions with probability $p_{t}$ proportional to the last encountered absolute temporal difference error,

\begin{equation}
    p_{t} \propto | r_{t+1} + \gamma_{t+1} \max_{u^{\prime}} Q_{\bar{\theta}}(s_{t+1}, u^{\prime}) - Q_{\theta}(s_{t}, u_{t}) |^{\omega},
\end{equation}

where $\omega$ is a tuneable hyperparameter for shaping the probability distribution. New transitions are added to the replay buffer with maximum priority to ensure all experiences will be sampled at least once to have their errors evaluated.

\secInLine{n-Step Q-learning}
Traditional Q-learning uses the target network's greedy action at the next step to bootstrap a Q-value estimate for the temporal difference target. Alternatively, to improve learning speeds and help with convergence \citep{Sutton1998, hessel2017rainbow}, forward-view \textit{multi-step} targets can be used \citep{Sutton1988}, where the $n$-step discounted return from state $s$ is

\begin{equation}
    r_{t}^{(n)} = \sum_{k=0}^{n-1} \gamma_{t}^{(k)} r_{t+k+1},
\end{equation}

resulting in an $n$-step DQN loss of 

\begin{equation}
    J_{DQN_{n}}(Q) = \big[ r^{(n)}_{t} + \gamma^{(n)}_{t} \max_{u^{\prime}} Q_{\bar{\theta}}(s_{t+n}, u^{\prime}) - Q_{\theta}(s_{t}, u_{t}) \big]^{2}.
\end{equation}


\subsection{Training Time and Convergence}
\label{app:training_time_and_convergence}

To train our \ac{rl} agent, we had a compute budget limited to one A100 GPU which was shared by other researchers from different groups. This resulted in highly variable training times. On average, one epoch on the large $500\times1000$ set covering instances took roughly $0.42$ seconds (which includes the time to act in the \ac{bnb} environment to collect and save the experience transitions, sample from the buffer, make online vs. target network predictions, update the network, etc.). Therefore training for $200\text{k}$ epochs (roughly the amount needed to converge on a strong policy within $\approx20\%$ of the imitation agent) took \num{5}-\num{6} days. 

\begin{figure}[tp!]
    \centering
    \includegraphics[width=0.5\textwidth]{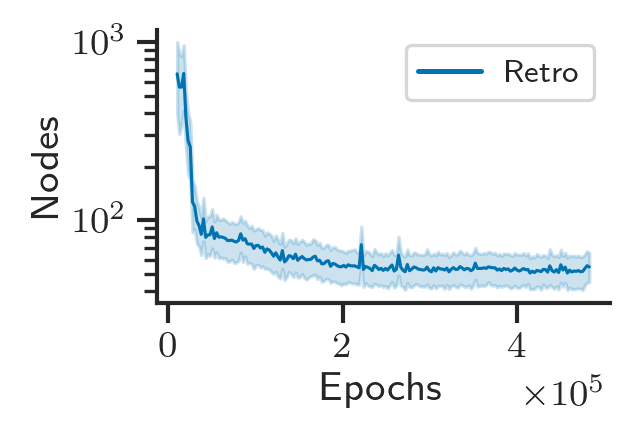}
    \caption[width=0.5\textwidth]{Validation curve for the retro branching agent on the $500\times1000$ set covering test instances. Although most performance gains were made in the first $\approx200\text{k}$ epochs, the agent did not stop improving, with the last recorded checkpoint improvement at $485\text{k}$ epochs.}
    \label{fig:rl_validation_curve_did_not_converge}
\end{figure}

As shown in Figure \ref{fig:rl_validation_curve_did_not_converge}, when we left our retro branching agent to train for $\approx13$ days ($\approx500\text{k}$ epochs), although most performance gains had been made in the first $\approx200\text{k}$ epochs, the agent never stopped improving (the last improved checkpoint was at $485\text{k}$ epochs). A potentially promising next step might therefore be to increase the compute budget of our experiments by distributing retro branching across multiple GPUs and CPUs and see whether or not the agent does eventually match or exceed the $500\times1000$ set covering performance of the \ac{il} agent after enough epochs.

\section{Neural Network}

\subsection{Architecture}
\label{app:network_architecture}

\begin{figure*}[tp!]
    \centering
    \includegraphics[width=0.9\textwidth]{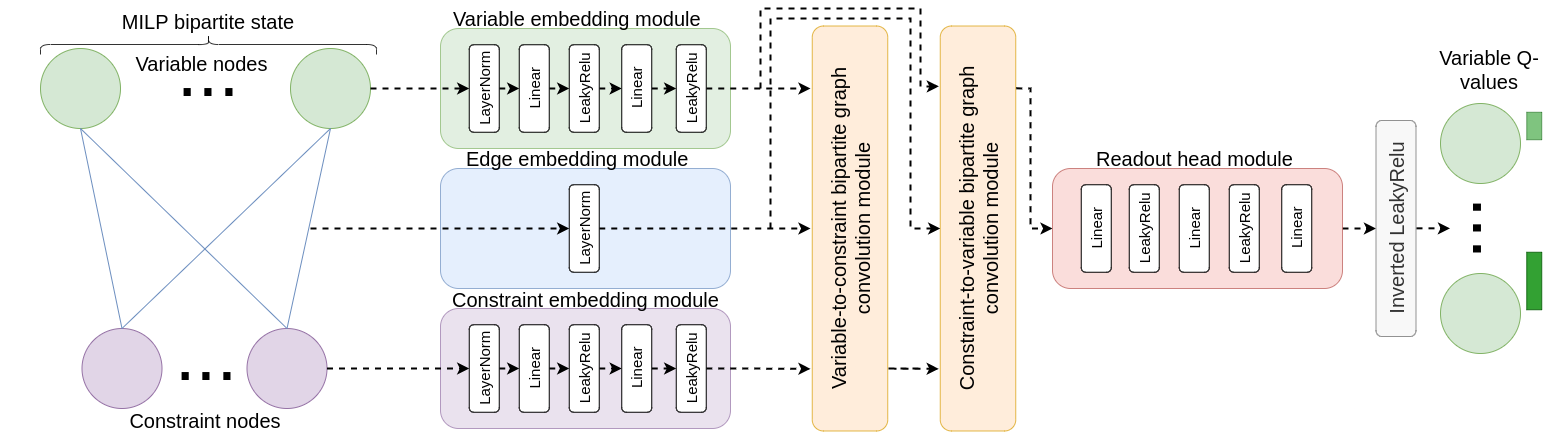}
    \caption[width=0.9\textwidth]{Neural network architecture used to parameterise the Q-value function for our ML agents, taking in a bipartite graph representation of the MILP and outputting the predicted Q-values for each variable in the MILP.}
    \label{fig:gnn_architecture}
\end{figure*}

We used the same \ac{gcn} architecture as \citealt{gasse2019} to parameterise our \ac{dqn} value function with some minor modifications which we found to be helpful. Firstly, we replaced the ReLU activations with Leaky ReLUs which we inverted in the final readout layer in order to predict the negative Q-values of our \ac{mdp}. 
Secondly, we initialised our linear layer weights and biases with a normal distribution ($\mu=0, \sigma=0.01$) and all-zeros respectively, and our layer normalisation weights and biases with all-ones and all-zeros respectively. Thirdly, we removed a network forward pass in the bipartite graph convolution message passing operation which we found to be unhelpfully computationally expensive. For clarity, Figure \ref{fig:gnn_architecture} shows the high-level overview of the neural network architecture. \ins{For a full analysis of the benefit of using \acp{gcn} for learning to branch, refer to \citealt{gasse2019}.}

\ins{

\subsection{Inference \& Solving Times}

The key performance criterion to optimise for any branching method is the reduction of the overall \ac{bnb} solving time. However, accurate and precise solving time and primal-dual integral over time comparisons are difficult because they are hardware-dependent. This is particularly problematic in research settings where CPU/GPU resources are often shared between multiple researchers and therefore hardware performance (and consequently solving time) significantly varies. Consequently, as in other works \cite{khalil2016LearningTB, gasse2019, etheve2020}, we presented and optimised for the number of \ac{bnb} tree nodes as this is hardware-independent and, in the context of prior work, can be used to infer the solving time. 

Specifically, we use the same \ac{gcn}-based architecture of \citealt{gasse2019} for all \ac{ml} branchers, thus all \ac{ml} approaches have the same per-step inference cost. Therefore the relative difference in the number of tree nodes is exactly the relative wall-clock times on equal hardware. When the per-step inference process is different (as for our non-\ac{ml} baselines, such as \ac{sb}), the number of tree nodes is not an adequate proxy for solving time. However, \citealt{gasse2019} have already demonstrated that the \ac{gcn}-based branching policies of \ac{il} outperform the solving time of other branchers such as \ac{sb}. As this \ac{ml} speed-up has already been established, in this manuscript we focus on improving the per-step \ac{ml} decision quality using \ac{rl} rather than further optimising network architecture, or otherwise, for speed, which we leave to further work.

However, empirical solving times are of interest to the broader optimisation community. Therefore, Table \ref{tab:branching_inferred_solving_times} provides a summary of the solving times of the branching agents on the large $500\times1000$ set covering instances under the assumption that they were ran on the same hardware as \citealt{gasse2019}.

}

\begin{table}[!htp]
    \centering
    \begin{tabular}{l | r}
    \toprule
         Method & Solving time (s) \\
         
         \midrule
         
         SB & \num{33.5} \\
         IL & \num{2.1} \\
         Retro & \num{2.5} \\
         FMSTS-DFS & \num{12.2} \\
         FMSTS & \num{7.6} \\
         Original & \num{35.8}
         
    \end{tabular}
    \caption{\ins{Inferred mean solving times of the branching agents on the large $500\times1000$ set covering instances under the assumption that they were ran on the same hardware as \citealt{gasse2019}.}}
    \label{tab:branching_inferred_solving_times}
\end{table}


\ins{

\section{Data Set Size Analysis}
\label{app:data_set_size_analysis}

As described in Section \ref{sec:experimental_setup}, we used \num{100} \ac{milp} instances unseen during training to evaluate the performance of each branching agent. This is in line with prior works such as \citealt{khalil2016LearningTB} who used \num{84} instances and \citealt{gasse2019} who used \num{20}. To ensure that \num{100} instances are a large enough data set to reliably compare branching agents, we also ran the agents on \num{1000} large $500\times1000$ set covering instances. The relative performance of each branching agent was approximately the same as when evaluated on \num{100} instances, with Retro scoring \num{65.3} nodes, FMSTS \num{250} ($3.8\times$ worse than Retro), \ac{il} \num{55.4} ($17.8\%$ better than Retro), and \ac{sb} \num{43.3}. In the interest of saving evaluation time and hardware demands and to make the development of and comparison to our work by future research projects more accessible, as well as for clarity in the per-instance Retro-IL comparison of Figure \ref{fig:500x1000_branching_performance_plots}d, we report the results for \num{100} evaluation instances in the main paper in the knowledge that the relative performances are unchanged as we scale the data set to a larger size.

}

\section{SCIP Parameters}
\label{app:scip_parameters}

For all non-\ac{dfs} branching agents we used the same \citealt{scip2020} \ac{bnb} parameters as \citealt{gasse2019}, as summarised in Table \ref{tab:scip_parameters}.

\begin{table}[!htp]
    \centering
    \begin{tabular}{lr}
    \toprule
         SCIP Parameter & Value \\
         
         \midrule
         
         \texttt{separating/maxrounds} & \num{0} \\
         \texttt{separating/maxroundsroot} & \num{0} \\
         \texttt{limits/time} & \num{3600}
         
    \end{tabular}
    \caption{Summary of the \citealt{scip2020} hyperparameters used for all non-\ac{dfs} branching agents (any parameters not specified were the default \citealt{scip2020} values).}
    \label{tab:scip_parameters}
\end{table}

\section{Observation Features}
\label{app:observation_features}
We found it useful to add \num{20} features to the variable nodes in the bipartite graph in addition to the \num{19} features used by \citealt{gasse2019}. These additional features are given in Table \ref{tab:observation_features}; their purpose was to help the agent to learn to aggregate over the uncertainty in the future primal-dual bound evolution caused by the partially observable activity occurring in sub-trees external to its retrospectively constructed trajectory.

\begin{table*}[!htp]
    \centering
    \begin{tabular}{lr}
    \toprule
         Variable Feature & Description \\
         
         \midrule
         
         \texttt{db\_frac\_change} & Fractional dual bound change  \\
         \texttt{pb\_frac\_change} & Fractional primal bound change \\
         \texttt{max\_db\_frac\_change} & Maximum possible fractional dual change \\
         \texttt{max\_pb\_frac\_change} & Maximum possible fractional primal change \\
         \texttt{gap\_frac} & Fraction primal-dual gap \\
         \texttt{num\_leaves\_frac} & \# leaves divided by \# nodes \\
         \texttt{num\_feasible\_leaves\_frac} & \# feasible leaves divided by \# nodes \\
         \texttt{num\_infeasible\_leaves\_frac} & \# infeasible leaves divided by \# nodes \\
         \texttt{num\_lp\_iterations\_frac} & \# nodes divded by \# LP iterations \\
         \texttt{num\_siblings\_frac} & Focus node's \# siblings divided by \# nodes \\
         \texttt{is\_curr\_node\_best} & If focus node is incumbent \\
         \texttt{is\_curr\_node\_parent\_best} & If focus node's parent is incumbent \\
         \texttt{curr\_node\_depth} & Focus node depth \\
         \texttt{curr\_node\_db\_rel\_init\_db} & Initial dual divided by focus' dual \\
         \texttt{curr\_node\_db\_rel\_global\_db} & Global dual divided by focus' dual \\
         \texttt{is\_best\_sibling\_none} & If focus node has a sibling \\
         \texttt{is\_best\_sibling\_best\_node} & If focus node's sibling is incumbent \\
         \texttt{best\_sibling\_db\_rel\_init\_db} & Initial dual divided by sibling's dual \\
         \texttt{best\_sibling\_db\_rel\_global\_db} & Global dual divided by sibling's dual \\
         \texttt{best\_sibling\_db\_rel\_curr\_node\_db} & Sibling's dual divided by focus' dual \\
         
    \end{tabular}
    \caption{Descriptions of the \num{20} variable features we included in our observation in addition to the \num{19} features used by \citealt{gasse2019}.}
    \label{tab:observation_features}
\end{table*}

\section{FMSTS Implementation}
\label{app:fmsts_implementation}
\citet{etheve2020} did not open-source any code, used the paid commercial \citet{cplex2009v12} solver, and experimented with proprietary data sets. Furthermore, they omitted comparisons to any other \ac{ml} baseline such as \citet{gasse2019}, further limiting their comparability. However, we have done a `best effort' implementation of the relatively simple \ac{fmsts} algorithm, whose core idea is to set the Q-function of a \ac{dqn} agent as minimising the sub-tree size rooted at the current node and to use a \ac{dfs} node selection heuristic. To replicate the \ac{dfs} setting of \citet{etheve2020} in \citet{scip2020}, we used the parameters shown in Table \ref{tab:dfs_scip_parameters}. We will release the full re-implementation to the community along with our own code. 

\begin{table*}[!htp]
    \centering
    \begin{tabular}{lr}
    \toprule
         SCIP Parameter & Value \\
         
         \midrule
         
         \texttt{separating/maxrounds} & \num{0} \\
         \texttt{separating/maxroundsroot} & \num{0} \\
         \texttt{limits/time} & \num{3600} \\
         
         \texttt{nodeselection/dfs/stdpriority} & \num{1073741823} \\
         \texttt{nodeselection/dfs/memsavepriority} & \num{536870911} \\
         
         \texttt{nodeselection/restartdfs/stdpriority} & \num{-536870912} \\
         \texttt{nodeselection/restartdfs/memsavepriority} & \num{-536870912} \\
         \texttt{nodeselection/restartdfs/selectbestfreq} & \num{0} \\
         
         \texttt{nodeselection/bfs/stdpriority} & \num{-536870912} \\
         \texttt{nodeselection/bfs/memsavepriority} & \num{-536870912} \\
         
         \texttt{nodeselection/breadthfirst/stdpriority} & \num{-536870912} \\
         \texttt{nodeselection/breadthfirst/memsavepriority} & \num{-536870912} \\
         
         \texttt{nodeselection/estimate/stdpriority} & \num{-536870912} \\
         \texttt{nodeselection/estimate/memsavepriority} & \num{-536870912} \\
         
         \texttt{nodeselection/hybridestim/stdpriority} & \num{-536870912} \\
         \texttt{nodeselection/hybridestim/memsavepriority} & \num{-536870912} \\
         
         \texttt{nodeselection/uct/stdpriority} & \num{-536870912} \\
         \texttt{nodeselection/uct/memsavepriority} & \num{-536870912} \\

    \end{tabular}
    \caption{Summary of the \citealt{scip2020} hyperparameters used the \ac{dfs} \ac{fmsts} branching agent of \citealt{etheve2020} (any parameters not specified were the default \citealt{scip2020} values).}
    \label{tab:dfs_scip_parameters}
\end{table*}

\section{Pseudocode}

\subsubsection{Retrospective Trajectory Construction}

Algorithm \ref{alg:subtree_episode_construction_pseudocode} shows the proposed `retrospective trajectory construction' method, whereby fathomed leaf nodes not yet added to a trajectory are selected as the brancher's terminal states and paths to them are iteratively established using some construction method.

\begin{algorithm}[!htp]
   \caption{Retrospectively construct trajectories.}
   \label{alg:subtree_episode_construction_pseudocode}
   
    \begin{algorithmic}
    
       \STATE \textbf{Input:} \ac{bnb} tree $\mathcal{T}$ from solving MILP
       \STATE \textbf{Output:} Retrospectively constructed trajectories
       
      
        \STATE \textbf{Initialise:} nodes\_added, subtree\_episodes = [$\mathcal{T}_{\text{root}-1}$], []
        
        \STATE // Construct trajectories until all valid node(s) in $\mathcal{T}$ added
        \WHILE{True}
        
            \STATE // Root trajectories at highest level unselected node(s)
            \STATE subtrees = []
            \FOR{node in nodes\_added}
                \FOR{child\_node in $\mathcal{T_{\text{node}}}\text{.children}$}
                    \IF{child\_node not in nodes\_added}
                        \STATE // Use depth-first-search to get sub-tree
                        \STATE subtrees.append(dfs($\mathcal{T}$, root=child\_node))
                    \ENDIF
                \ENDFOR
            \ENDFOR
            
            \STATE // Construct trajectory episode(s) from sub-tree(s)
            \IF{$\text{len(subtrees)} > 0$}
                \FOR{subtree in subtrees}
                    \STATE subtree\_episode = construct\_path(subtree) (\ref{alg:subtree_path_construction_pseudocode})
                    \STATE subtree\_episode[$-1$].done = True 
                    \STATE subtree\_episodes.append(subtree\_episode)
                    \FOR{node in subtree\_episode}
                        \STATE nodes\_added.append(node)
                    \ENDFOR
                \ENDFOR
            
            \ELSE
                \STATE // All valid nodes in $\mathcal{T}$ added to a trajectory
                \STATE \textbf{break}
            
            \ENDIF
            
        \ENDWHILE
        
      
    \end{algorithmic}

\end{algorithm}

\subsubsection{Maximum Leaf \Ac{lp} Gain}

Algorithm \ref{alg:subtree_path_construction_pseudocode} shows the proposed `maximum leaf \ac{lp} gain' trajectory construction method, whereby the fathomed leaf node with the greatest change in the dual bound (`\ac{lp} gain') is used as the terminal state of the trajectory.

\begin{algorithm}[!htp]
   \caption{Maximum leaf LP gain trajectory construction.}
   \label{alg:subtree_path_construction_pseudocode}
   
    \begin{algorithmic}
    
       \STATE \textbf{Input:} Sub-tree $\mathcal{S}$
       \STATE \textbf{Output:} Trajectory $\mathcal{S}_{E}$
       \STATE \textbf{Initialise:} gains = \{\}
       
       \FOR{leaf in $\mathcal{S}\text{.leaves}$}
        \IF{leaf closed by brancher}
            \STATE gains.leaf = $| \mathcal{S}_{\text{root}}\text{.dual\_bound} - \mathcal{S}_{\text{leaf}}\text{.dual\_bound} |$
        \ENDIF
       \ENDFOR
       
       \STATE terminal\_node = max(gains)
       
       \STATE $\mathcal{S}_{E}$ = shortest\_path(source=$\mathcal{S}_{\text{root}}$, target=terminal\_node)
      
    \end{algorithmic}

\end{algorithm}

\section{Cost of Strong Branching Labels}
\label{app:cost_strong_branching_labels}

\begin{figure}
    \centering
    \includegraphics[width=0.45\textwidth]{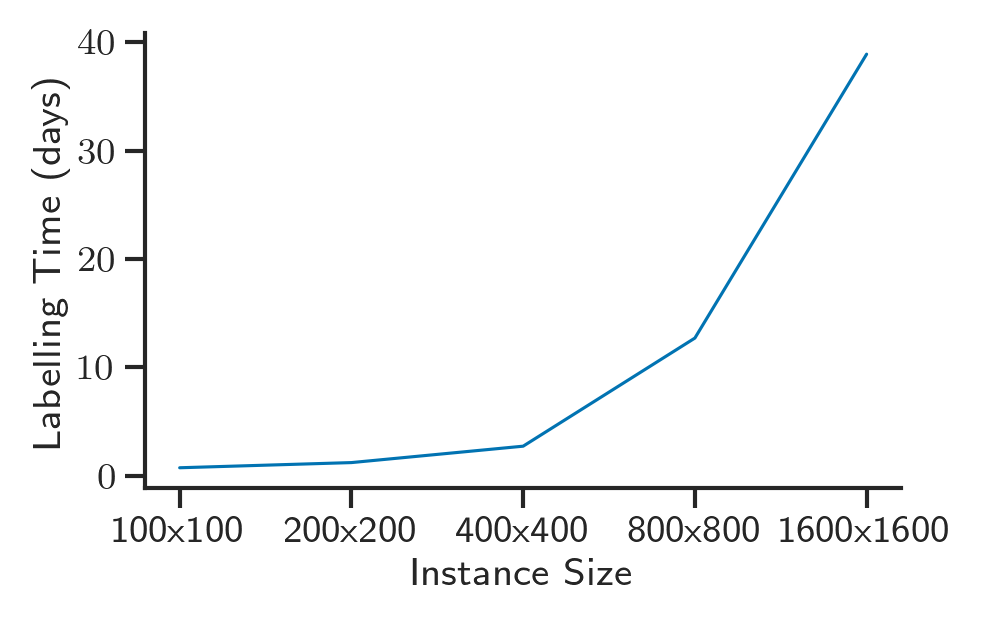}
    \caption{How the explore-then-strong-branch data labelling phase of the strong branching imitation agent scales with set covering instance size ($\text{rows} \times \text{columns}$) using an Intel Xeon ES-2660 CPU and assuming \num{120000} samples are needed for each set.}
    \label{fig:explore_then_strong_branch_data_labelling_scaling}
\end{figure}

As well as performance being limited to that of the expert imitated, \ac{il} methods have the additional drawback of requiring an expensive data labelling phase. Figure \ref{fig:explore_then_strong_branch_data_labelling_scaling} shows how the explore-then-strong-branch labelling scheme of \citealt{gasse2019} scales with set covering instance size ($\text{rows} \times {\text{columns}}$) and how this becomes a hindrance for larger instances. Although an elaborate infrastructure can be developed to try to label large instances at scale \citep{nair2021solving}, ideally the need for this should be avoided; a key motivator for using \ac{rl} to branch.

\end{document}